\documentclass[10pt,twocolumn]{article}
\usepackage[a4paper,margin=18mm,columnsep=6mm]{geometry}
\usepackage{mathptmx}
\usepackage[T1]{fontenc}
\usepackage{graphicx,booktabs,amsmath,microtype}
\usepackage[font=small,labelfont=bf]{caption}
\usepackage[hidelinks]{hyperref}
\hypersetup{pdftitle={Using AI to Simulate the Behaviour of European Populations},pdfauthor={Umair Mehmood Imam}}

\title{\Large\bfseries Using AI to Simulate the Behaviour of European Populations}
\author{Umair Mehmood Imam\\[2pt]\normalsize Bineric AI, Oslo, Norway\\\normalsize support@bineric.com \quad ORCID 0009-0002-7892-9189}
\date{November 2026}

\begin{document}
\twocolumn[
\maketitle
\begin{center}\begin{minipage}{0.86\textwidth}
\textbf{Abstract.} Language models can stand in for survey respondents, but leading approaches build their
simulated people from interviews or individual records. We show that a model can learn how European
population groups respond from published statistics, without interviews and without training on individual
records. We fine-tune an open-weight language model on more than 12 million published group-level answer
distributions from about 40 European countries, setting it up to predict each group's \emph{departure} from a
reference distribution. On opinion surveys from its training families it calls the direction of group
differences right 84\% of the time, ahead of a demographic statistical model (77\%, paired test) and level with
Claude Opus 5 and 5.5, which it also ties on accuracy; the frontier models are more accurate on new questions.
On 278 held-out questions, almost all from the survey families seen in training, its error is 12\% lower than that
statistical model. It runs on a single 32\,GB GPU or a high-memory laptop, and in a test with an earlier model, an evidence gate calibrated on half of a survey's questions fell back to the reference where the model's adjustments did not help. Coupled to a transport simulation of a planned metro closure, with a conventional
elasticity as the reference, it adds a testable age gradient to the response, which we freeze for later scoring.

\end{minipage}\end{center}\vspace{1.2em}]

\section{Introduction}
Decisions about public services are made before anyone knows how people will respond. A transit
authority closing a metro section, a city changing fares, or a ministry redesigning a benefit all need an
estimate of how different groups will react. Surveys answer such questions, but slowly and at a cost, and
rarely for the exact group and change at hand.

Language models offer a shortcut. Prompted with a description of a person, they produce answers that
resemble those of real respondents \cite{argyle2023}. Several lines of work have pushed this further: agents
built from two-hour interviews with 1,052 people \cite{park2024}; models fine-tuned on 2.9 million individual
responses from social-science experiments \cite{kolluri2025}; and models fine-tuned on published opinion
distributions of subpopulations \cite{suh2025,santurkar2023,cao2025}. Interview-based approaches depend on records of identifiable individuals, which in Europe are personal data and, when they touch health, politics or religion, special-category data. Most of this work is built on US populations, and in cross-national tests the respondent's country acts as an
assumption that further respondent detail does not revise \cite{wang2026}.

We develop this model as the behaviour component of Simulayer (\url{https://simulayer.ai}), a platform that
simulates how people behave and move when something changes, such as a new service, a closure or a new price: a
behaviour model predicts how each population group responds, and a movement model carries the resulting trips
through a city's transport network. We study the European, privacy-preserving end of this design space. Our
contributions are:
\begin{enumerate}\itemsep0pt
\item \textbf{Behaviour from published statistics.} A model trained on published group-level results from about 40
countries, without interviews and without individual records, that predicts each group's departure from a
reference distribution (Sec.~\ref{sec:method}).
\item \textbf{Group differences at frontier level.} On opinion surveys from its training families it reads the
direction of group differences as well as two frontier models and better than a statistical model, and on held-out
questions from those families it is 12\% more accurate than the statistical model (Sec.~\ref{sec:results}, Table~\ref{tab:headline}).
\item \textbf{Deployable and safe.} It runs on one 32\,GB GPU or a high-memory laptop, and, in a test with an earlier model, an evidence gate calibrated per survey fell back to published figures where the adjustments did not help
(Secs.~\ref{sec:deploy},~\ref{sec:gate}).
\item \textbf{Ready for simulation.} The reference can be any conventional estimate, so the model plugs into
existing planning tools; we show this on a transport simulation of a metro closure (Sec.~\ref{sec:app}).
\item \textbf{A sealed evaluation protocol} with held-out releases, a baseline fixed in advance and intervals
clustered by question (Sec.~\ref{sec:eval}).
\end{enumerate}

\begin{table*}[t]\centering\small
\begin{tabular}{@{}lrr@{}}\toprule
Result & Ours & Comparison\\\midrule
Direction of group differences right$^{b}$ & 84.3\% & 82.9\% / 80.9\%$^{a}$; 76.8\%$^{c}$\\
Accuracy, opinion surveys$^{b}$ (TVD) & 0.0608 & 0.0616 / 0.0607$^{a}$\\
Accuracy, new questions (TVD) & 0.0468 & 0.0369 / 0.0331$^{a}$\\
Held-out questions$^{d}$ vs statistical model (TVD) & 0.0449 & 0.0509$^{c}$ ($-12\%$)\\
Held-out questions$^{d}$ vs national figure (TVD) & 0.0449 & 0.0556 ($-19\%$)\\
Runs on & 32\,GB GPU / laptop & provider API$^{a}$\\
Interviews or individual training records & none & --\\\bottomrule
\end{tabular}
\caption{Headline results. $^{a}$Claude Opus 5 / Opus 5.5 (few-shot). $^{b}$400 opinion-survey cells from the training families; an unmoved answer counts as half right; paired differences in Sec.~\ref{sec:frontier}. $^{c}$Demographic statistical model.
$^{d}$From survey families seen in training.}
\label{tab:headline}\end{table*}

\section{Related work}
\textbf{Simulating respondents.} Argyle et al.\ introduced ``silicon samples'' \cite{argyle2023}; later work
documented distorted opinion distributions and exaggerated effects in prompted models \cite{bisbee2024}, and
benchmarks of distributional alignment \cite{meister2024}.
\textbf{Agent architectures.} Generative agents with memory and reflection \cite{park2023} and populated
prototypes of social systems \cite{park2022} established the simulation paradigm; interview-based agents reached
85\% of participants' own test--retest consistency on the General Social Survey \cite{park2024}.
\textbf{Fine-tuning on human responses.} SubPOP fine-tunes on response distributions of US subpopulations
\cite{suh2025}; Cao et al.\ fine-tune on first-token answer probabilities for global populations \cite{cao2025};
SOCRATES fine-tunes on individual experimental responses and, in its main results, improves distributional
alignment on unseen studies by 26\% over its base model \cite{kolluri2025}; Centaur does the same for cognitive
experiments \cite{binz2025}. These works compare mainly against other language models and simple guesses. We
add an explicit statistical baseline fixed before training and an evidence gate for unfamiliar surveys.
\textbf{Behaviour in transport simulation.} Language-model agents have been used to choose travel modes inside
traffic simulators \cite{lammer2026}; we instead keep a conventional reference and let the model adjust it.

\section{Data}\label{sec:data}
\textbf{Sources.} The corpus consists of results that public producers have already published at group level:
opinion surveys, official statistics, reported-behaviour surveys, choice experiments and a small amount of
observed behaviour. It covers about 40 countries, more than 12 million group-level answer distributions and
more than 10,000 distinct questions. Groups are defined by the publishers (for example age band, sex,
education, region, or two-way combinations). All sources are public. We did not use interviews and did not train on individual
records; where a publisher released only anonymised respondent files, we used group shares computed from
them. No answer distribution in the training data was produced by a model. A share of training batches used openly licensed general conversation text, which contains no answer distributions. We do not name the sources; counts
and categories are given so the scope can be judged.

\textbf{Splits.} Test sets come from releases newer than any training data, and a development set from the
release before. For the main test sets, training rows whose question matched a test or development question by wording were removed before training. A separate set of \emph{new questions} contains questions whose wording never appears on the
training side.

\section{Method}\label{sec:method}
\textbf{Task.} A \emph{cell} is one group's answer distribution for one question. Let $q$ be a question with
$K$ answer options, $g$ a population group, $p_g\in\Delta^{K-1}$ the group's published distribution and
$p_0\in\Delta^{K-1}$ a reference distribution, by default the national result for the same question. The model
is a map
\begin{equation}
\hat p_g = F_\theta(g, q, p_0) \in \Delta^{K-1},
\end{equation}
trained so that $\hat p_g$ is close to $p_g$ over all training cells.

\begin{figure*}[t]\centering
\includegraphics[width=0.78\textwidth]{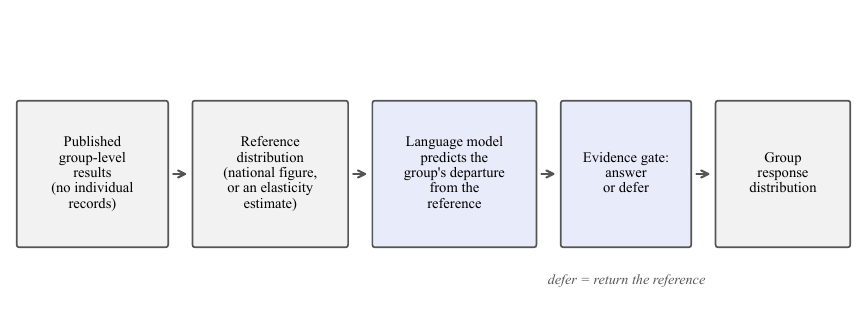}
\caption{The method. Given a reference distribution, the model predicts a group's departure from it, and an evidence gate decides whether that departure is used.}
\label{fig:method}\end{figure*}

\textbf{Predicting departures.} $F_\theta$ is constructed so that it reduces to the reference whenever the learned adjustment is zero,
$F_\theta(g,q,p_0)=p_0$, so that training concentrates on how a group departs from the reference
(Fig.~\ref{fig:method}).
The reference can be any distribution: in the application (Sec.~\ref{sec:app}) it is a conventional
elasticity estimate.

\textbf{Training.} We fine-tune an open-weight language model of a few tens of billions of parameters on
the published shares. We report no further training details. We also report an earlier checkpoint trained
the same way on less data.

\section{Evaluation protocol}\label{sec:eval}
\textbf{Metric.} Total variation distance between predicted and published distributions,
\begin{equation}
\mathrm{TVD}(p_g,\hat p_g)=\tfrac12\textstyle\sum_{k=1}^{K}\lvert p_{g,k}-\hat p_{g,k}\rvert ,
\end{equation}
the share of answers that would have to move to another option to match the published result. Lower is
better; we report the mean over cells.

\textbf{Baselines.} (i) the national figure; (ii) carry-forward: the national figure plus the group's last published gap from it; (iii) a demographic statistical model, fixed before training as the bar to beat, that adds the group's typical departure from its national figure, $\bar\delta_g$, learned only from earlier
questions with the same answer format (per country and group, shrunk towards the pooled average), clips at
zero and renormalises:
\begin{equation}
\hat p^{\,\mathrm{demo}}_{g,k}\propto\max\bigl(0,\;p_{0,k}+\bar\delta_{g,k}\bigr).
\end{equation} We also compare the earlier checkpoint and two frontier models given the
same input.

\textbf{Uncertainty.} Cells from one question are correlated, so 95\% intervals resample whole questions
(2,000 bootstrap draws; 5,000 for the unpaired frontier accuracy tables). 

\textbf{Direction agreement.} Let $S$ be the (cell, option) pairs whose published departure from the national
figure exceeds sampling noise for the cell's sample size $n_c$,
$\lvert p_{c,k}-p_{0,c,k}\rvert>1.96\sqrt{p_{0,c,k}(1-p_{0,c,k})/n_c}$. Then
\begin{equation}
\mathrm{DA}=\frac{1}{\lvert S\rvert}\sum_{(c,k)\in S}\mathbf 1\!\left[\operatorname{sgn}(\hat p_{c,k}-p_{0,c,k})=\operatorname{sgn}(p_{c,k}-p_{0,c,k})\right].
\end{equation}

\textbf{Exposure.} Most held-out releases were fielded before the base model was released, so exposure of the
base model to them in pre-training cannot be excluded; one test set of 2,145 cells was fielded after
(Sec.~\ref{sec:postcut}).

\section{Results}\label{sec:results}
\subsection{Group differences at frontier level}\label{sec:frontier}
\begin{figure*}[t]\centering
\includegraphics[width=0.92\textwidth]{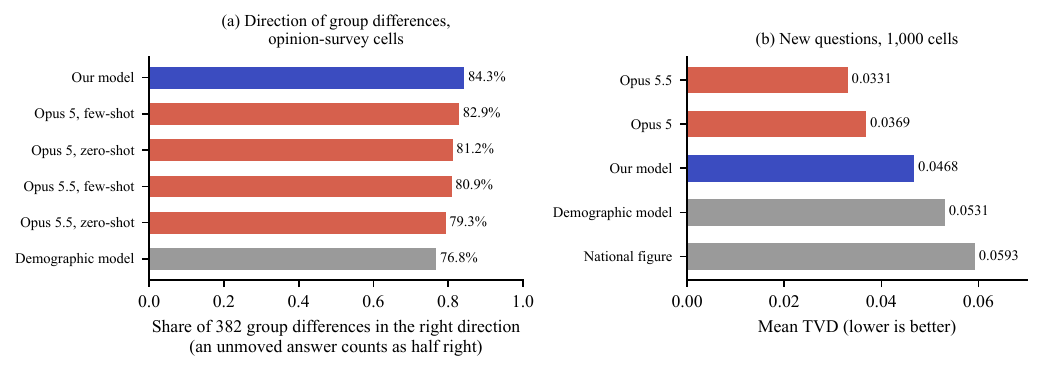}
\caption{Comparison with frontier models. (a) Direction of group differences on 400 opinion-survey cells,
on the same 382 group differences, an unmoved answer counted as half right; our model is level with the frontier models and ahead of the statistical model. (b) Accuracy on 1,000 new-question cells.}
\label{fig:frontier}\end{figure*}

Two frontier models (Claude Opus 5 and 5.5) received the same input through a commercial API. On 400 random
cells from the newest opinion-survey releases there are 382 group differences larger than sampling noise. Our model
calls the direction of 84.3\% of them right (Fig.~\ref{fig:frontier}a). The frontier models answer in whole
percentages and leave 14--20\% of these differences exactly at the national figure; counting such an unmoved answer
as half right, they get 79.3--82.9\%, and the paired differences from our model ($+1.4$ to $+5.0$ percentage points)
have 95\% intervals that include zero or, for Opus 5.5 zero-shot, touch it. Where each model moves, both are right 87--88\% of the time. Our model is ahead of the
demographic statistical model, which gets 76.8\% (paired difference $+7.5$ [$+2.8$, $+12.4$]). Accuracy on these
cells is a statistical tie with the frontier models (TVD 0.0608 against 0.0607--0.0624) and clearly better than the
statistical model (0.0676). On new questions the frontier models are more accurate (0.0331 and 0.0369 against
0.0468; Fig.~\ref{fig:frontier}b); they are general-purpose models served through an API, while ours runs on the
user's own hardware.

\subsection{Held-out questions}\label{sec:newq}
\begin{figure}[t]\centering
\includegraphics[width=\columnwidth]{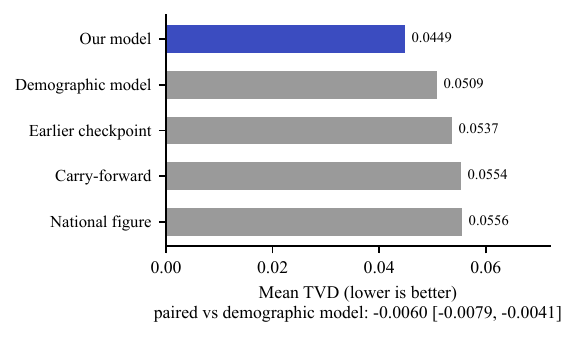}
\caption{Held-out new questions (278 questions, 114,729 cells). Whiskers: 95\% interval clustered by question.}
\label{fig:newq}\end{figure}

On 278 held-out new questions (114,729 cells) the model's mean TVD is 0.0449 [0.0378, 0.0552], against
0.0509 for the demographic model, 0.0554 for carry-forward and 0.0556 for the national
figure (Fig.~\ref{fig:newq}). The paired difference from the demographic model is $-0.0060$
[$-0.0079$, $-0.0041$], 12\% lower error. The earlier checkpoint does not beat the bar (0.0537). The model is
better than the bar on 168 of 278 questions, which, apart from one migration-attitudes study, come from survey
families seen in training.

\textbf{Releases fielded after the base model.}\label{sec:postcut} On 2,145 cells fielded after the base model's
release, our model beats the statistical model (0.0458 against 0.0496) and the national figure (0.0506). These are
later releases of mostly familiar questions (85\% of rows repeat a question asked in earlier waves), so the set
measures tracking over time rather than new questions; carrying the last gap forward (0.0405) and the frontier
models (0.0372--0.0388) are more accurate on it.

\subsection{Opinion surveys}
On the newest releases of the opinion surveys in the training corpus (183,544 cells, 167 questions) the
model's TVD is 0.0650 against 0.0719 for the national figure, 9.7\% lower. The earlier checkpoint is slightly
better here (0.0642; difference 0.0008 [0.0006, 0.0009]).

\subsection{\texorpdfstring{Runs on one 32\,GB GPU or a high-memory laptop}{Runs on one 32 GB GPU or a high-memory laptop}}\label{sec:deploy}
A compressed version that fits on a single 32\,GB GPU loses no measurable accuracy ($+0.0004$
[$-0.0005$, $+0.0014$] TVD on new questions, $+0.0013$ [$-0.0001$, $+0.0027$] on opinion surveys). A version for a high-memory laptop gives nearly the same answers as the compressed GPU version (mean difference 0.0033 TVD per cell; 0.0621 against 0.0622 on 400 cells) in about 0.3\,s per cell.

\subsection{An evidence gate for unfamiliar surveys}\label{sec:gate}\label{sec:transfer}
Group adjustments hold within the survey families the model was trained on. On a survey the model has never seen, the system starts from
the reference and an evidence gate decides when to use the model's adjustment. The gate is a decision rule
\begin{equation}
\tilde p_g=\begin{cases}\hat p_g & \text{if } d_s\,\sigma(\hat p_g,p_0)\ge\tau_s,\\ p_0 & \text{otherwise,}\end{cases}
\end{equation}
where $\sigma$ is a simple signal computed from the model's own output, and the threshold $\tau_s$ and the
orientation $d_s\in\{-1,+1\}$ are fitted per survey $s$ on calibration questions, because the signal's meaning
differs between surveys. In a test with a model from before both checkpoints reported here, on test sets built before our sealed protocol, calibrated on half of a survey's questions and scored on the other half, the gate kept the full gain on opinion surveys (0.0683
against 0.0778 for the reference, answering 74\% of cells) and, on a transit operator's survey where the model's
adjustments did not help, returned the reference (0.0513, answering 2\%) instead of losing accuracy (0.0624
ungated) (Fig.~\ref{fig:gate}). Results outside the main training families are in Appendix~\ref{app:transfer}.

\begin{figure}[t]\centering
\includegraphics[width=0.95\columnwidth]{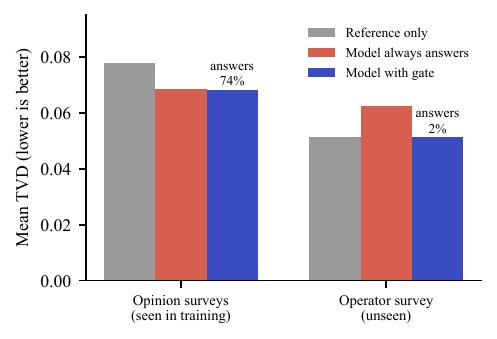}
\caption{The evidence gate (a model from before both checkpoints), calibrated on half of each survey's questions and scored on the rest.}
\label{fig:gate}\end{figure}

\section{Application: a metro closure}\label{sec:app}
\begin{figure*}[t]\centering
\includegraphics[width=0.92\textwidth]{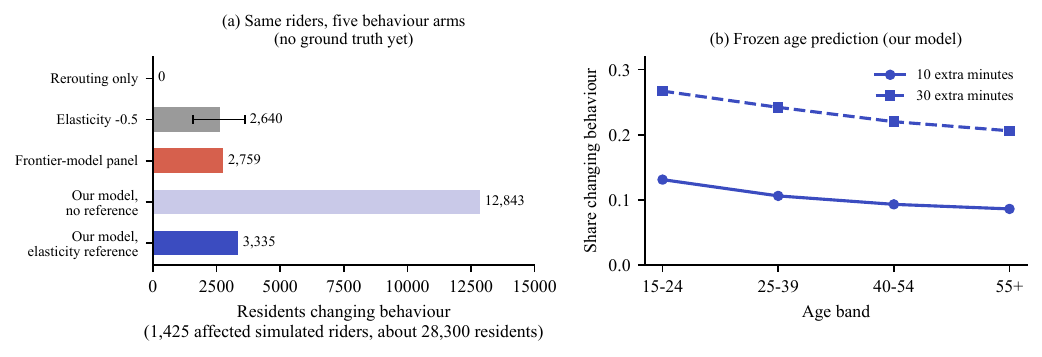}
\caption{Metro closure, an integration test without ground truth. (a) The same 1,425 affected riders under five
behaviour arms (whisker: elasticity $-0.3$ to $-0.7$). (b) Our model's frozen prediction: share changing
behaviour by age band and extra time.}
\label{fig:closure}\end{figure*}

This section shows the model inside Simulayer's movement simulation. It is a preliminary application and an integration test: the closure has not happened, so nothing here can be scored. We consider a planned months-long closure of a metro section, replaced by buses. In our own agent-based transport simulation of the region, built on a synthetic population, 1,425 affected simulated public-transport riders (representing about 28,300 residents) experience extra travel time. Holding the network, population and each rider's
extra time fixed, we redrew only the behavioural response under five arms (Fig.~\ref{fig:closure}a, residents changing behaviour): rerouting only; a conventional travel-time elasticity of $-0.5$; a table produced earlier by a frontier model answering
for a synthetic resident panel; our model asked without a reference; and our model given the elasticity as its
reference. Traffic was not re-simulated after the responses, so the loop in which changed behaviour alters crowding and travel times and, in turn, later decisions is not yet closed; closing it is the next milestone. Given the elasticity as its reference, as the design intends, 11.8\% of riders change behaviour, against 9.3\% for the elasticity alone: 7\% drive or make fewer trips and 5\% shift their travel time, rising with the delay;
and adds a group dimension the elasticity lacks: riders under 40 change behaviour more than older riders
(Fig.~\ref{fig:closure}b). The reference is what makes this work: asked without one, the model's answers barely
respond to the delay and 45\% of riders change behaviour. How
the elasticity's losses split between driving, fewer trips and other times is our assumption, not the model's.
We recorded the SHA-256 hash of this prediction and its scoring rule before seeing any outcome data
(\texttt{\small 5ec1bf7b9021ddff5a14ec366ffabc75}\-\texttt{\small 3a21054cde09a4524b8a6d42128d4f0e}), to be scored against observed behaviour.

\section{Discussion}
\textbf{Privacy by design.} The model learns from published group results, without interviews or training on individual records, so no consent process or special-category data is involved.
\textbf{European evidence.} It reflects how European groups answer, from European sources, rather than
transferring US behaviour.
\textbf{Deployable.} It runs on a single 32\,GB GPU or a high-memory laptop inside the user's own infrastructure.
\textbf{Complements existing models.} Because the reference can be any estimate, a city's own model output or a
standard elasticity, it adds group differences to the tools planners already trust instead of replacing them,
and the gate, so far tested on an earlier model, falls back to those tools where the model has no evidence.
\textbf{General.} The same approach applies wherever statistics are published by group: health, energy,
finance, mobility.

\section{Limitations}
The gains apply to survey families represented in training; frontier models are more accurate on new questions,
and on the one test set fielded after the base model's release a carry-forward baseline is more accurate. The
training data are mostly survey answers rather than responses to interventions, and the gate needs some history
from each survey. We report one training run, and because sources and training details are not disclosed the
results cannot be reproduced exactly; the protocol and baselines are described so the claims can be checked
against any comparable corpus.

\section{Ethics and data statement}
No interviews were conducted and no individual records were used as training data; a minority of group shares
were computed from anonymised public respondent files. No individual can be identified from the model's
outputs, which are group shares. Group-level predictions can still reinforce stereotypes if read as facts about
individuals; the gate is designed to fall back to published figures where the model is unreliable. Frontier-model outputs were used only as comparison baselines, never as training data.

\section{Conclusion}
A language model trained on published statistics alone can learn how European population groups depart from their
country's answer. On opinion surveys from its training families it reads the direction of group differences as well as Claude Opus 5 and 5.5 and better than a statistical model, and on held-out questions it beats that statistical model. It runs on one 32\,GB GPU, needs no interviews or individual records, and in a test with an earlier model an evidence gate kept that model from making estimates worse where its adjustments did not help. Because it adjusts any reference estimate, it can add group differences to the
planning tools public authorities already use, as our metro-closure example shows.

\section*{Statements}
\textbf{Competing interests.} The author is the founder of Bineric AI (Bineric AS), which develops Simulayer, a commercial platform based on the model described here. The model weights and training corpus are proprietary.
\textbf{Funding.} This work received no external research funding.
\textbf{Data availability.} The training corpus is not released; its sources are public and available from their
producers under their published terms. The test-only datasets include the European Social Survey and the PACES
migration-attitudes study, available from their producers.
\textbf{Code availability.} Model weights are not released. Scoring code is available from the author on request.
\textbf{Use of AI assistants.} Language-model assistants (Claude by Anthropic and ChatGPT by OpenAI) helped with
code, analysis scripts and drafting. The author checked every number against the result files and is responsible
for the content.

\appendix
\section{Results outside the main training families}\label{app:transfer}
\begin{table}[h]\centering\small
\begin{tabular}{@{}p{3.6cm}rrr@{}}\toprule
Source & Cells & National & Our model\\\midrule
Social attitudes survey & 3,095 & 0.0795 & 0.0815\\
Transit operator's passenger survey & 507 & 0.0573 & 0.0607\\
Migration attitudes study & 3,820 & 0.0422 & 0.0508\\
Travel surveys (11 questions)$^{a}$ & 159 & 0.101 & 0.108\\\bottomrule
\end{tabular}
\caption{Mean TVD (lower is better). $^{a}$Earlier results of these surveys were in training in small amounts; difference $+0.007$
[$-0.002$, $+0.018$], direction of group differences right in 56\% of 605 pairs. The migration study is also
part of the 278 new questions in Sec.~\ref{sec:newq}.}
\end{table}
\end{document}